\documentclass[letterpaper, 10 pt, conference]{ieeeconf}  

\IEEEoverridecommandlockouts                              

\usepackage{balance}
\usepackage[noadjust]{cite}
\usepackage{xcolor}
\usepackage{amssymb}
\usepackage{amsfonts}
\usepackage{amsmath}
\usepackage{wrapfig}
\usepackage{booktabs}
\usepackage{graphicx}
\usepackage{multirow}
\usepackage{float}
\usepackage{array}
\usepackage{hyperref}
\usepackage[flushleft]{threeparttable}

\newcommand{\etal}{\textit{et al}.}

\newcommand{\eg}{\textit{e}.\textit{g}.,}

\newcommand{\overbar}[1]{\mkern 1.5mu\overline{\mkern-1.5mu#1\mkern-1.5mu}\mkern 1.5mu}

\newcolumntype{N}{>{\centering\arraybackslash}m{.38in}}
\newcolumntype{G}{>{\centering\arraybackslash}m{.3in}}

\title{\LARGE \bf
Towards Real-Time Generation of Delay-Compensated Video Feeds for Outdoor Mobile Robot Teleoperation
}

\author{Neeloy Chakraborty*$^{1}$, Yixiao Fang*$^{1}$, Andre Schreiber$^{1}$, Tianchen Ji$^{1}$, Zhe Huang$^{1}$, Aganze Mihigo$^{2}$,\\Cassidy Wall$^{3}$, Abdulrahman Almana$^{1}$, and Katherine Driggs-Campbell$^{1}$%
\thanks{*Denotes equal contribution. Authors affiliated with the departments of $^{1}$Electrical and Computer Engineering, $^{2}$Computer Science, and $^{3}$Industrial and Enterprise Systems Engineering at the University of Illinois Urbana-Champaign. Emails: \{neeloyc2, yixiaof2, andrems2, tj12, zheh4, amihigo2, cwall20, aalmana2, krdc\}@illinois.edu}%
}

\begin{document}

\maketitle
\thispagestyle{empty}
\pagestyle{empty}


\begin{abstract}

Teleoperation is an important technology to enable supervisors to control agricultural robots remotely.
However, environmental factors in dense crop rows and limitations in network infrastructure hinder the reliability of data streamed to teleoperators.
These issues result in delayed and variable frame rate video feeds that often deviate significantly from the robot's actual viewpoint.
We propose a modular learning-based vision pipeline to generate delay-compensated images in real-time for supervisors. 
Our extensive offline evaluations demonstrate that our method generates more accurate images compared to state-of-the-art approaches in our setting.
Additionally, ours is one of the few works to evaluate a delay-compensation method in outdoor field environments with complex terrain on data from a real robot in real-time.
Resulting videos and code are provided at {\color{cyan}{\href{https://sites.google.com/illinois.edu/comp-teleop}{\textit{https://sites.google.com/illinois.edu/comp-teleop}}}}.

\end{abstract}


\section{Introduction}

\begin{figure}[t]
    \centering
    \includegraphics[width=\linewidth]{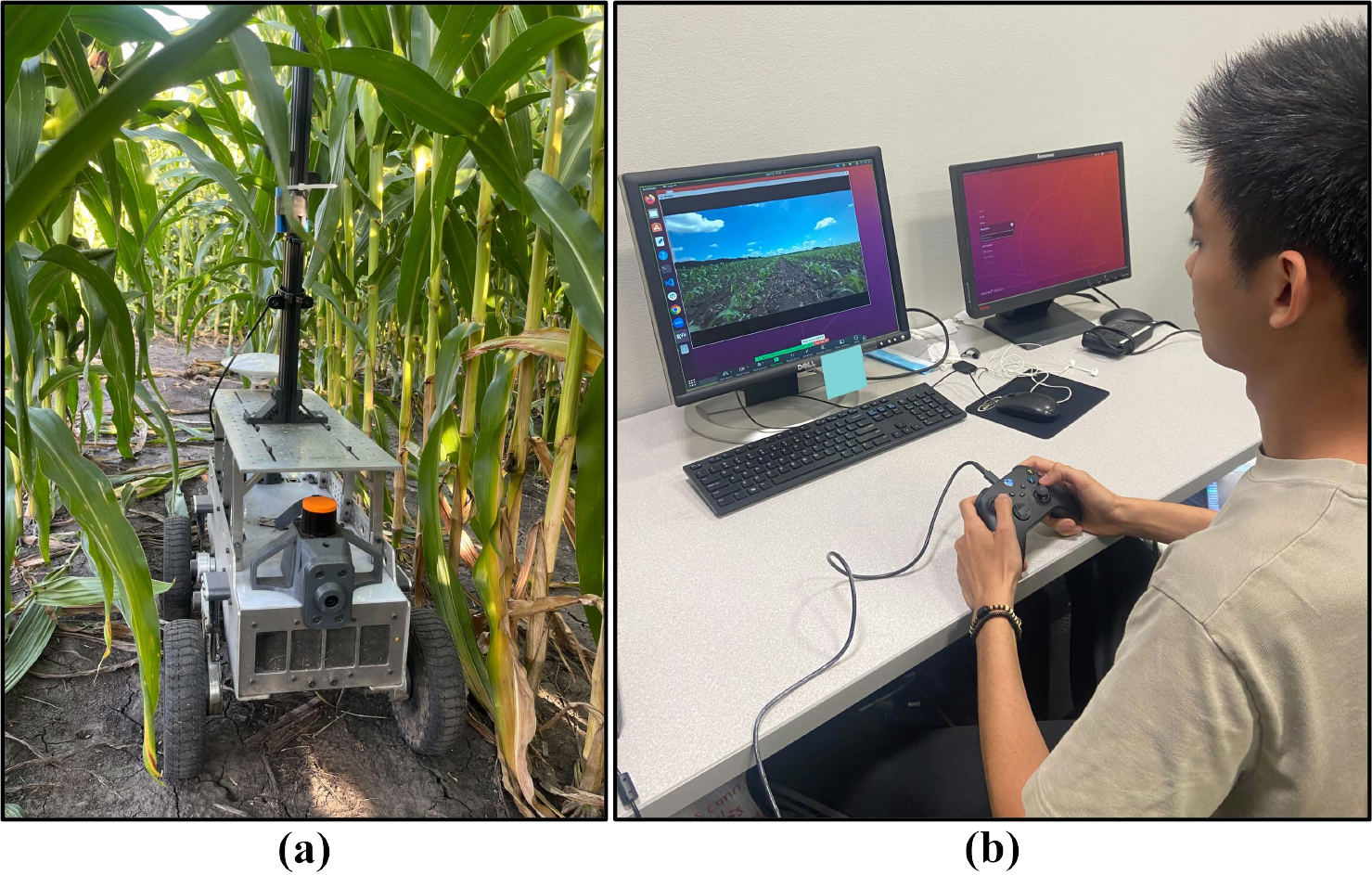}
    \vspace{-20pt}
    \caption{\textbf{(a) The TerraSentia+ robot in dense growth and (b) an example of a remote teleoperation setup.}}
    \label{fig:terra_cover}
    \vspace{-15pt}
\end{figure}

Robots are finding their way into increasingly complex application areas, spanning manufacturing, healthcare, security, entertainment, and other industries~\cite{shen2021robots}.
One such field that has growing interest is the agriculture domain, where mobile robots are used for phenotyping, enriching soil, predicting yield, and more~\cite{droukas2023survey}.
Although their autonomous capabilities have drastically improved in recent years~\cite{xu2022review,sivakumar2021learned,ji2022proactive,gasparino2024wayfaster}, there still exist instances where it is preferred that a human supervisor manually control the robot remotely via \emph{teleoperation}.
During teleoperation, a supervisor views a stream of data (\eg~video, pose) sent from the robot and sends action commands through a remote controller, as shown in Figure~\ref{fig:terra_cover}.
In our case, a supervisor is needed to visually inspect crops and manually control the robot when the autonomy stack fails. 

While teleoperation is a necessary technology for these robots, we find through real-world testing that our organization's existing teleoperation platform\footnote[4]{Details of the hardware behind the networking setup of the teleoperation system are discussed by Sie~\etal~\cite{sie2025byon}, and is out of the scope of this work.} has several limitations, including severe delay in communication between the robot and supervisor, and intermittent transmission failures~\cite{sie2025byon}.
These transmission issues are caused by inherent delay in sending information over a network with low bandwidth, and moisture in the crops causing signal fades in parts of the farm.
Furthermore, the challenging under-canopy crop environment and unpredictable weather patterns lead to uneven terrain, causing large random deviations in consecutive camera poses.
In extreme cases, severe delays may lead to undesired robot crashes, causing catastrophic task failure.

While we cannot remove all instances of delays and frame skips in the teleoperation pipeline, we can compensate for missing information at any time by predicting the frame to be shown to the supervisor.
This task, known as \emph{frame delay compensation} or \emph{apparent latency reduction}, is well studied for indoor robots~\cite{farajiparvar2020survey,moniruzzaman2022teleoperation,darvish2023teleoperation}.
Although researchers have tackled this problem in several domains, to the best of our knowledge, few works study the effect of delay compensation approaches on mobile robots deployed in outdoor scenarios~\cite{chen2020real,zheng2020evaluation,brudnak2016predictive,moniruzzaman2022high,moniruzzaman2023long,moniruzzaman2023structure,prakash2023predictive,lee2024latency}.
Even then, existing works each have their own limitations, including only predicting latency-compensated robot poses~\cite{zheng2020evaluation}, being evaluated solely in simulation~\cite{zheng2020evaluation,brudnak2016predictive,moniruzzaman2022high,moniruzzaman2023long,moniruzzaman2023structure}, relying on access to an RGB-D camera during inference~\cite{chen2020real,prakash2023predictive}, and using an over-simplified video prediction model with assumptions that cannot be transferred to under-canopy field robots~\cite{brudnak2016predictive,moniruzzaman2022high}.
Additionally, most methods have been tested in on-road driving environments with much more predictable terrain and camera motion compared to the under-canopy scenario~\cite{prakash2023predictive,lee2024latency}.

In contrast, our approach (1) predicts both robot poses and camera images, (2) is tested on data from a real robot in challenging field environments, (3) simply assumes access to a monocular camera alongside an estimate of the robot's pose, and (4) uses learning-based approaches to accurately and efficiently generate images.
In particular, our modular pipeline consists of a monocular metric depth estimation (MDE) model, a robot kinematics model, an efficient sphere-based renderer, and a learning-based inpainting model.
Notably, we showcase the feasibility of finetuning state-of-the-art depth estimation foundation models to our complex environment, with which we generate $3$D colored point clouds at runtime.
A simple robot kinematics model is used to predict future poses conditioned on user actions, which are passed into a renderer to generate images.
Finally, an inpainting model fills in holes in the rendering before the image is shown to the supervisor.
We find the work presented by Prakash~\etal~\cite{prakash2023predictive} is most related to our proposed method.
However, they rely on the availability of an RGB-D camera at runtime, their method is applied to the autonomous driving domain resulting in a simplified rendering approach, and the authors do not provide quantitative results analyzing the accuracy of generated images. 

Our primary contributions are summarized as follows: 
we 
(1) design an efficient modular pipeline for frame delay compensation, 
(2) extensively compare our pipeline with ablations, classical image processing methods, and state-of-the-art learning-based image generation approaches on an offline crop dataset in diverse growth stages, 
and (3) showcase the real-time operation of our approach on real-world data by integrating our method into a ROS node.

\section{Related Works}
\label{sec:related_works}

\subsection{Time Delay Compensation for Robot Teleoperation}

Teleoperation is a primary mode of control in robotics for challenging tasks and environments where full automation is still actively under development, like space exploration~\cite{sheridan1993space}, underwater operation~\cite{khatib2016ocean}, nuclear material handling~\cite{marturi2016towards}, and more~\cite{martins2009immersive,meng2004remote,zhang2020toward,murakami2008development}. 
In many of these applications, the human operator is expected to be far from the robot, introducing issues of limited bandwidth and perception latency~\cite{farajiparvar2020survey}. 
In particular, high latency results in serious consequences where real-time responsiveness is critical~\cite{liaw2017target}.
Different methodologies have been investigated to mitigate the impact of delays on teleoperation, including devising a move-and-wait user strategy for tasks which allow quasi-static operations~\cite{hokayem2006bilateral}, and abstracting away low-level short-term control signals with high-level long-term user commands~\cite{ghosh2020human}. 
Another avenue of research for delay compensation is future frame prediction~\cite{zheng2020evaluation}. 
By incorporating the operator control commands, a simple method that uses sliding and zooming video transformation for prediction can achieve impressive performance gain in driving scenarios~\cite{moniruzzaman2022high}. 
More recent efforts are harnessing neural networks to account for missing details from boundary disocclusions~\cite{moniruzzaman2023structure, moniruzzaman2023long, lee2024latency}. 
Note that there are very few works on delay compensation for field robots due to the complexity of required networking infrastructure and the unstructured nature of outdoor environments~\cite{chen2020real}.

\subsection{Video Prediction and View Synthesis}

We specifically choose to tackle latency compensation with video prediction methods.
Early research in video prediction focuses on deterministic models predicting in raw pixel space~\cite{lotter2016deep,gao2022simvp}, which requires expensive image reconstruction from scratch. 
To promote efficiency, later studies pivoted towards high-level predictions in feature space, such as optical flow in DMVFN~\cite{hu2023dynamic} and segmentation maps in S2S~\cite{luc2017predicting}. 
Such models often warp or inpaint input images based on predicted image features for video prediction. 
However, deterministic models inherently confine possible motion outcomes to a single, fixed result, resulting in blurry images~\cite{ming2024survey}. 
To overcome this issue, SRVP~\cite{franceschi2020stochastic}, a variational neural network, models the temporal evolution of the system through a latent state, which is conditioned on learned stochastic variables and is later transformed into predicted images. 
In the domain of view synthesis, which is the task of generating new views of a scene given one or more images, the most similar work to ours is SynSin~\cite{wiles2020synsin}. 
The end-to-end model constructs a $3$D point cloud of \textit{latent features}, which is then projected to the target view and inpainted to generate the output image. 
Neural rendering often plays an important role in such view synthesis algorithms~\cite{tewari2020state}. 
Typical neural renderers require a mesh-based geometry representation, which prohibits topology change and drags rendering speed~\cite{loper2014opendr,kato2018neural,petersen2019pix2vex}. 
As a result, our work incorporates a sphere-based renderer, Pulsar, which has been shown to have real-time capability~\cite{lassner2021pulsar}.

\subsection{Monocular Depth Estimation with Deep Learning}

Modular pipelines, like ours, may use a monocular depth estimation model to predict the depth of pixels in camera images.
Deep learning methods in particular have produced state-of-the-art results in this field.
A pioneering work proposed training a convolutional neural network (CNN) combining global and local predictions to produce a final depth output~\cite{eigen2014depth}. 
However, monocular MDE can be challenging due to varying scenes and sensors, and alternative methods for relative depth estimation (RDE) have been introduced~\cite{chen2016single, ranftl2020towards}. 
Later advancements also involved using the Vision Transformer~\cite{dosovitskiy2020image} as an encoder instead of a CNN~\cite{ranftl2021vision} to provide greater global image context and improved prediction accuracy.
However, training solely on specialized datasets can result in poor estimates when transferred to a new environment.
MiDaS~\cite{ranftl2020towards} attempts to overcome this problem for RDE by mixing various datasets together. 
ZoeDepth~\cite{bhat2023zoedepth} extends this idea to MDE by first pretraining a network on the relative depth estimation task, and then finetuning a subset of parameters on the MDE task with different datasets.
Alternatively, powerful visual foundation models, like DINOv2, can help mitigate the drop in performance caused by domain shift~\cite{oquab2024dinov2,yang2024depthv1,yang2024depth}.
Nonetheless, collecting high-quality real-world labeled depth data for tasks like finetuning remains challenging due to noise~\cite{yang2024depth}.

\begin{figure}[t]
    \centering
    \includegraphics[width=0.9\linewidth]{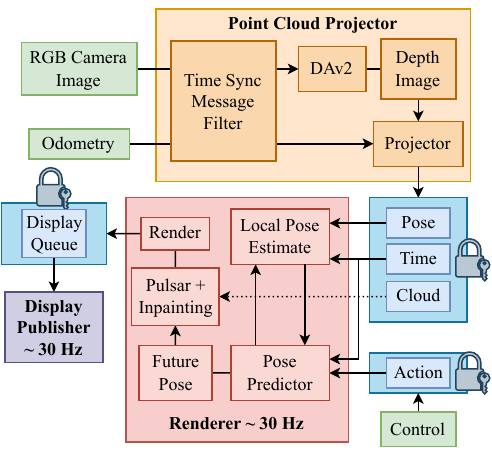}
    \vspace{-5pt}
    \caption{\textbf{Block diagram of ROS pipeline.}
             The robot sends sensor messages (green) to our node.
             Functions are required to wait for mutex locks when accessing or modifying global data (blue).
             The renderer generates images that are 30 Hz apart to enable a consistent FPS display.}
    \label{fig:ros_block}
    \vspace{-15pt}
\end{figure}

\section{Methods}
\label{sec:methods}

\subsection{Problem Formulation}

We formulate the apparent latency reduction problem as a video prediction task.
Specifically, given a sequence of $m+1$ video frames $I_{t-m:t}\in\mathbb{R}^{H\times W\times3}$ and poses $P_{t-m:t}\in\mathbb{R}^{4\times4}$ from time $t-m$ to $t$ from a camera with calibrated intrinsic matrix $K\in\mathbb{R}^{3\times3}$, we aim to predict $I_{t+1:t+n}$.

\subsection{Choosing a Scene Representation}
\label{sec:scene_rep}

One common approach to generating future frames conditioned on $I_{t-m:t}$, is to predict future $2$D pixel flows $F_{t+1:t+n}$, and iteratively apply each flow to $I_{t}$~\cite{liu2017video,liang2017dual,nagabhushan2022temporal}.
Although effective in environments with smooth camera motions, flow-based methods fail in deployments where consecutive input frames appear far apart and have few correspondences, as is the case in choppy low-bandwidth networks like our setting.

A simpler non-learning-based approach to predicting $I_{t+1:t+n}$ conditioned on just $I_{t}$, is to estimate the future camera poses $P_{t+1:t+n}$, optimize for homography matrices $H_{t+1:t+n}$, and apply each homography to $I_{t}$~\cite{brudnak2016predictive}.
While efficient, this method relies on a planar scene representation of the world where most of the scene in front of the camera is a flat plane.
However, in the case of scenes with large variance in object depths, camera motions will result in a range of projected pixel locations in the future image plane.
This phenomena is particularly prevalent in environments like cluttered crop rows, where closer crops will move larger distances in the image plane when the camera shifts. 

Rather than assuming the scene is a plane, given a depth map $D_{t}\in\mathbb{R}^{(H\cdot W)\times1\times1}$ of the distance to each pixel in $I_{t}$, we calculate the $3$D point cloud $y$ in the camera frame:
\[
y = K^{-1}XD\in\mathbb{R}^{(H\cdot W)\times3\times1},
\]
where $X\in\mathbb{N}^{(H\cdot W)\times3\times1}$ is the homogeneous representation of the pixel coordinates in $I_{t}$.
Then, with an estimate of future camera extrinsics $\overline{P}$, we compute the unnormalized projected pixel coordinates $\widetilde{X}$ in the future images:
\[
\widetilde{X} = \overline{K}{\overbar{P}}^{-1}PY\in\mathbb{R}^{(H\cdot W)\times3\times1};\quad \overline{K} = \left[K|0\right]\in\mathbb{R}^{3\times4},
\]
where $Y\in\mathbb{R}^{(H\cdot W)\times4\times1}$ is the homogeneous representation of $y$.
Finally, we normalize $\widetilde{X}$ by the new depth of each point to compute the projected homogeneous coordinates $\overline{X}$. 

Point cloud representations are versatile and common in the vision community~\cite{kivi2022real,wegen2024survey,kato2020differentiable}.
However, directly applying the projection process above leads to several holes in the rendered image where camera motion uncovered occlusions.
While $3$D Gaussian Splatting~\cite{kerbl2023gaussian} is a promising method for rendering point clouds efficiently, the training time to optimize gaussian parameters for describing a scene in high-fidelity is still not real-time~\cite{lee2024compact,matsuki2024gaussian,girish2023eagles,jin2024lighting}.
Instead, using the efficient renderer discussed in Section~\ref{sec:methods_pulsar}, we represent our scene as a set of spheres, each with its own radius and blending weight simply determined by their distances from the camera, intrinsic parameters, and rasterization settings.

\subsection{Depth Estimation}
\label{sec:depth_methods}

Before we can create a point cloud to render images from, we need an estimate of the depth image $D$ from $I$.
RGB-D stereo cameras enable an accurate measurement of $D$, but the teleoperated robot may not have such a sensor installed.
Furthermore, from experiments, we have found our sensor's (ZED 2) depth measurement is noisy and has several holes outdoors, which results in unknown pixels in the rendered images. 
As such, we draw on the recent advancements in depth estimation foundation models, and finetune the Depth Anything V2 (DAv2)~\cite{yang2024depth,yang2024depthv1} weights to our environment.
Given an input image $I$, DAv2 attempts to minimize the root scale-invariant loss~\cite{eigen2014depth} between the predicted depth $\widetilde{d}^{(i)}$ and label $d^{(i)}$ at each pixel $i\in1\ldots N$:
\[
\mathcal{L}_{\text{depth}} = \sqrt{\frac{1}{N}\sum_{i=1}^Nd_{\log}(i)^2 - \frac{\lambda}{N^2}\left(\sum_{i=1}^Nd_{\log}(i)\right)^2},
\]
where $d_{\log}(i) = \log{d^{(i)}} - \log{\widetilde{d}^{(i)}}$, and $\lambda\in\mathbb{R}_{\geq0}$ is a parameter to balance the accuracy and sharpness of predictions.

While DAv2 generates a large set of synthetic labels from simulators, existing mobile robot crop row simulations are too low-fidelity to curate an informative dataset from~\cite{pinto2023navigating}.
Instead, we randomly select a set of videos from a recent real-world under-canopy mobile robot dataset collected by Cuaran~\etal~\cite{cuaran2024under}, which includes ground truth depth from a ZED 2 camera onboard a TerraSentia, and we finetune DAv2-Small on a subset of images and pixels with known depth.
Although the labels are noisy, we find DAv2 is capable of transferring to our environment.
Further details about the dataset and training procedure are provided in Section~\ref{sec:offline_setup}.

\subsection{Future Pose Prediction}

During real-world deployment, we need predictions of the future trajectory of the robot so as to render subsequent frames.
Thus, we rely on a simple skid-steer kinematic model to estimate the $x$ and $y$ position of the robot, as well as its heading $\theta$, given a linear $v$ and angular $\omega$ velocity command:
\[
\dot{x} = \mu v\cos(\theta);\quad \dot{y} = \mu v\sin(\theta);\quad \dot{\theta} = \eta\omega,
\]
where $\mu$ and $\eta$ are friction coefficients.
Gasparino~\etal~\cite{gasparino2022wayfast} predict these coefficients with a learned neural network.
To improve runtime efficiency and reduce memory usage, we set $\mu=\eta=1$ during evaluation, effectively simplifying the kinematics to the Dubins' car model~\cite{dubins1957curves}.
We note, however, that $\mu$ and $\eta$ may be tuned or adaptively chosen in real-time for the specific terrain.

\begin{figure}[t]
    \centering
    \includegraphics[width=\linewidth]{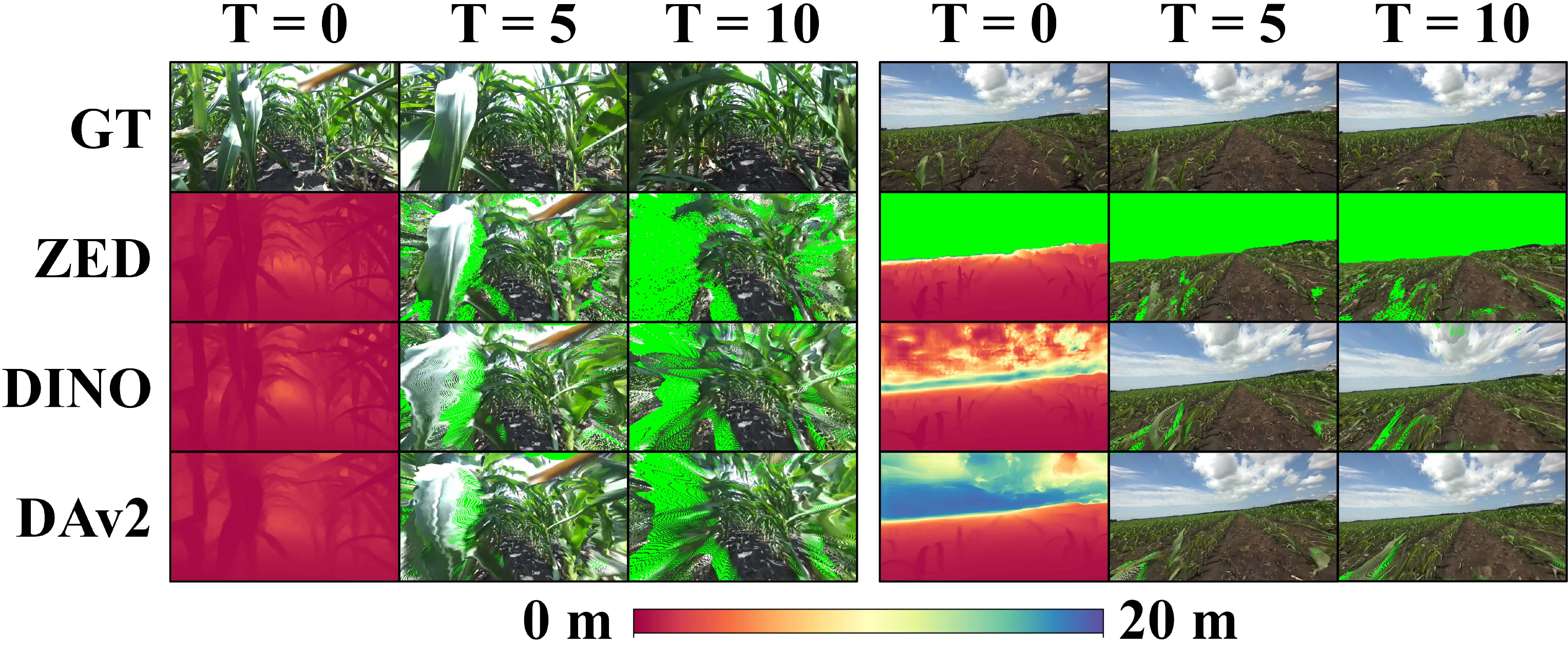}
    \vspace{-15pt}
    \caption{\textbf{A comparison of depth model estimates and resulting Pulsar renderings with ground truth (GT).}
             Holes in predictions are drawn green.}
    \label{fig:depth_comparison}
    \vspace{-10pt}
\end{figure}

\subsection{Rendering with Pulsar}
\label{sec:methods_pulsar}

Based on our discussions from Section~\ref{sec:scene_rep}, we choose to represent the environment as a set of colored spheres with some opacity, and we use Pulsar~\cite{lassner2021pulsar} to render novel scenes in real-time.
Pulsar sets the radius $r^{(i)}$ of each sphere $i$ in the scene dynamically according to the distance between the camera and sphere, normalized device coordinate (NDC) intrinsics, and a chosen rasterization radius constant $R$:
\[
r^{(i)} = \frac{RK_w\left\Vert P-p^{(i)}\right\Vert_2}{2K_f},
\]
where $K_w$ and $K_f$ are the camera's sensor width and focal length respectively.
Intuitively, points closer to the camera are given smaller radii during rendering.
Unlike the sampling procedure of early neural radiance field approaches~\cite{mildenhall2020nerf}, Pulsar only stores memory for occupied regions in space, enabling efficient computation of the rendered color for a pixel.
In particular, Pulsar uses a softmax blending function to weight overlapping points along a ray when rendering the projected image (Eq. 1 from Lassner~\etal~\cite{lassner2021pulsar}).
Increasing the $\gamma$ parameter in this function enables us to raise the transparency of nearby points.
Finally, and arguably most important for real-world deployment, Pulsar is integrated into PyTorch~\cite{ansel2024pytorch}, eliminating the bottleneck of transferring DAv2 predictions or point clouds between libraries or devices --- required for other renderers~\cite{schutz2022software,schutz2021rendering} --- and it is implemented with specialized CUDA kernels that allow for real-time rendering of images on our hardware.

\subsection{Inpainting Network}
\label{sec:methods_inpainting}

Camera motion to the predicted future pose will lead to holes in rendered images output from Pulsar, even with its sphere representation. 
As such, like SynSin~\cite{wiles2020synsin}, we learn a neural network to refine rendered images.
Specifically, after rendering a raw future image $\widetilde{I}_{t+k}$, we first embed the original image $I_t$ and $\widetilde{I}_{t+k}$ through a ResNet-18 encoder pretrained by Sivakumar~\etal~\cite{sivakumar2021learned} on real-world TerraSentia navigation data, storing intermediate hidden states for later skip connections.
We then concatenate both embeddings and intermediate hidden states through a series of decoder blocks consisting of convolution, batch normalization, ReLU, and upsampling layers to reach the original resolution of $I_t$.
The inpainting model $g$ is trained to minimize mean absolute error between the prediction and the ground truth $I_{t+k}$:
\[
\mathcal{L}_{\text{refine}} = \frac{1}{N}\sum_{i=1}^N\left\vert I_{t+k}^{(i)} - g\left(I_t, f\left(I_t, D_t, P_t, P_{t+k}\right)\right)^{(i)}\right\vert,
\]
where the function $f$ outputs the raw rendered image $\widetilde{I}_{t+k}$ with Pulsar.
We find that after training, the refined images are blurry, but generally resemble the shapes of objects in the ground truth future frame.
As such, during inference, we only fill in holes in $\widetilde{I}_{t+k}$ with corresponding predictions from $g\left(I_t,\widetilde{I}_{t+k}\right)$, and display the final result to the supervisor.
Training details are provided in Section~\ref{sec:offline_setup}.

\section{Offline Evaluation}
\label{sec:offline_eval_whole}

\begin{table*}[t]
\centering
\caption{
\textbf{Average accuracy of metric depth models across different crop growth stages.}
Models predict depth in meters.
\\AbsRel is better when lower, while higher values of $\delta_1$, PSNR, and FPS are desired.
FPS is measured on a 2080 GPU.
}
\vspace{-5pt}
\label{tab:depth_acc}
\noindent
\resizebox{\textwidth}{!}{%
\begin{tabular}{c N N N N N N N N N N N N c}
\toprule
\multirow{2}{*}{Model}  & \multicolumn{3}{c }{Early ($45\%$ ZED Holes)} & \multicolumn{3}{c }{Middle ($0.036\%$ ZED Holes)} &\multicolumn{3}{c }{Late ($0.037\%$ ZED Holes)} & \multicolumn{3}{c }{Average ($14\%$ ZED Holes)} & \multirow{2}{*}{FPS}\\
\cmidrule(ll){2-4}\cmidrule(ll){5-7}\cmidrule(ll){8-10}\cmidrule(ll){11-13}
& AbsRel & $\delta_1$ & $\text{PSNR}_{t+5}$ & AbsRel & $\delta_1$ & $\text{PSNR}_{t+5}$ & AbsRel & $\delta_1$ & $\text{PSNR}_{t+5}$ & AbsRel & $\delta_1$ & $\text{PSNR}_{t+5}$ \\ 
\midrule
\multicolumn{1}{ l }{DINOv2-Terra} & $\textbf{0.101}$ & $\textbf{0.889}$ & $18.467$ & $\textbf{0.212}$ & $\textbf{0.681}$ & $\textbf{11.233}$ & $\textbf{0.242}$ & $\textbf{0.608}$ & $\textbf{12.149}$ & $\textbf{0.186}$ & $\textbf{0.722}$ & $13.901$ & $41$\\ 
\multicolumn{1}{ l }{DAv2-vKITTI} & $0.982$ & $0.076$ & --- & $1.007$ & $0.075$ & --- & $0.811$ & $0.170$ & --- & $0.929$ & $0.108$ & --- & $\textbf{58}$\\ 
\multicolumn{1}{ l }{DAv2-Terra} & $0.128$ & $0.783$ & $\textbf{18.848}$ & $0.289$ & $0.555$ & $11.044$ & $0.331$ & $0.500$ & $12.018$ & $0.251$ & $0.609$ & $\textbf{13.917}$ & $\textbf{58}$\\
\bottomrule
\end{tabular}
}
\vspace{-10pt}
\end{table*}

\subsection{Experimental Setup}
\label{sec:offline_setup}

\subsubsection*{\textbf{Dataset}}

We collect several data points from the public dataset provided by Cuaran~\etal~\cite{cuaran2024under} to train and evaluate models on the video prediction task.
Specifically, we extract synchronized $720$p images, depth maps, camera poses, and action inputs at $16$ Hz from several rosbags in different crop growth stages. 
Our processed dataset contains $7382$ training and $1847$ validation labels across $2$ early-, $3$ middle-, and $4$ late-stage growth videos, while the test split holds $3100$ labels from $1$ video per growth stage.
The sequences are collected from a calibrated ZED 2 camera with neural depth set to output a range between $0.2$ and $20$ meters.

\subsubsection*{\textbf{Training}}

We finetune the DAv2-Small model pretrained on the Virtual KITTI 2~\cite{cabon2020virtual} synthetic dataset for $120$ epochs with a batch size of $32$ and learning rates of $5\mathrm{e}{-5}$ and $5\mathrm{e}{-4}$ for the DINOv2~\cite{oquab2024dinov2} encoder and DPT~\cite{ranftl2021vision} head weights respectively.
Images and labels are resized to $518\times518$ before training and we minimize the scale-invariant loss against known ZED neural depth values.
Then, we collect an augmented dataset to train the inpainting model.
Using the finetuned depth model and Pulsar, we render projected $720$p training images at $t\mathrel{+}=\{1,3,5,7\}$ given ground truth future poses.
Pulsar is configured to store $1$ sphere of $R=3\mathrm{e}{-3}$ per pixel, and $\gamma=0.1$.
Holes, or disocclusions, are rendered as green.
The inpainting model is trained to minimize $\mathcal{L}_{\text{refine}}$ using the ground truth future images as labels for $100$ epochs with a batch size of $16$ and learning rate of $5\mathrm{e}{-4}$.

\begin{table}[t]
\centering
\caption{
\textbf{Image inpainting quality of different models.}
\\All three metrics are better when higher.
}
\vspace{-5pt}
\label{tab:inpainting}
\resizebox{\linewidth}{!}{%
\begin{tabular}{m{0.4in}m{0.3in}ccc}
\toprule
\multicolumn{1}{c}{Metric}  & \multicolumn{1}{c}{Delay}     & Telea~\cite{telea2004image}   & ResNet-L1         & ResNet-MS-SSIM    \\ 
\midrule
\multirow{2}{*}{PSNR}       & $t+5$                         & $14.278$                      & $\textbf{14.377}$ & $14.351$          \\
                            & $t+10$                        & $13.482$                      & $\textbf{13.937}$ & $13.853$          \\
\midrule
\multirow{2}{*}{MS-SSIM}    & $t+5$                         & $\textbf{0.343}$              & $0.342$           & $0.342$           \\
                            & $t+10$                        & $0.285$                       & $\textbf{0.287}$  & $0.285$           \\
\midrule
\multicolumn{2}{l}{FPS}                                     & $0.386$                       & $\textbf{36}$     & $\textbf{36}$     \\
\bottomrule
\end{tabular}
}
\vspace{-15pt}
\end{table}

\subsubsection*{\textbf{Baselines}}

We compare our pipeline against a variety of other real-time video prediction and latency-compensation approaches, including:
(1) a non-learning-based approach presented by Moniruzzaman~\etal~\cite{moniruzzaman2022high} that predicts a cropped window within $I_t$ conditioned on robot state and action input, and returns an upsampled version of the window to the user (denoted as C$+$S for crop and scale); 
(2) SRVP~\cite{franceschi2020stochastic}, a variational neural network trained to learn a distribution of latent states, which are sampled to generate future images; 
(3) DMVFN~\cite{hu2023dynamic}, a state-of-the-art flow-based video prediction model;
and (4), the end-to-end novel view-synthesis model SynSin~\cite{wiles2020synsin}, which renders images from a point cloud of latent features.
To predict longer horizon sequences from SRVP and DMVFN, we iteratively use intermediate predictions to generate future frames.
During offline evaluation, we assume SynSin and our pipeline have access to ground truth future poses for rendering, and we apply our pose prediction model on real-time data in Section~\ref{sec:online_eval_whole}.

\subsection{Results}
\label{sec:offline_results}

\begin{table*}[t]
\centering
\caption{
\textbf{Generation quality of video prediction methods across crop growth stages and time delays on offline data.}
\\PSNR, MS-SSIM, and FPS are better when higher, while LPIPS is better when lower.
\underline{Underline} denotes second-best. 
}
\vspace{-5pt}
\label{tab:baseline_comparison}
\noindent
\resizebox{\textwidth}{!}{%
\begin{tabular}{l l G G G G G G G G G G G G c}
\toprule
\multicolumn{1}{c}{\multirow{2}{*}{Model}} & \multicolumn{1}{c}{\multirow{2}{*}{Metric}} & \multicolumn{3}{c }{Early} & \multicolumn{3}{c }{Middle} &\multicolumn{3}{c }{Late} & \multicolumn{3}{c }{Average} & \multirow{2}{*}{FPS}\\
\cmidrule(ll){3-5}\cmidrule(ll){6-8}\cmidrule(ll){9-11}\cmidrule(ll){12-14}
&& $t+1$ & $t+5$ & $t+10$ & $t+1$ & $t+5$ & $t+10$ & $t+1$ & $t+5$ & $t+10$ & $t+1$ & $t+5$ & $t+10$ \\ 
\midrule
\multirow{3}{*}{C$+$S~\cite{moniruzzaman2022high}}      & PSNR      & $15.010$                & $14.279$                & $13.913$                    
                                                                    & $10.327$                & $9.993 $                & $9.816 $                    
                                                                    & $11.195$                & $11.021$                & $10.914$                    
                                                                    & $12.151$                & $11.744$                & $11.531$                    
                                                                    & \multirow{3}{*}{$\underline{28}$}                                                       \\  
                                                        & MS-SSIM   & $0.292$                 & $0.276$                 & $0.271$                     
                                                                    & $0.125$                 & $0.107$                 & $0.097$                     
                                                                    & $0.146$                 & $0.137$                 & $0.133$                     
                                                                    & $0.186$                 & $0.172$                 & $0.166$                 \\  
                                                        & LPIPS     & $0.546$                 & $\underline{0.580}$     & $\underline{0.603}$         
                                                                    & $\underline{0.575}$     & $\underline{0.624}$     & $\underline{0.653}$         
                                                                    & $0.625$                 & $\underline{0.652}$     & $\underline{0.671}$         
                                                                    & $0.583$                 & $\underline{0.619}$     & $\underline{0.643}$     \\  
\cmidrule(ll){1-2}
\multirow{3}{*}{SRVP~\cite{franceschi2020stochastic}}   & PSNR      & $18.857$                & $16.504$                & $13.541$                    
                                                                    & $\underline{11.872}$    & $\textbf{11.760}$       & $\textbf{11.613}$           
                                                                    & $12.740$                & $\textbf{12.846}$       & $\textbf{12.616}$           
                                                                    & $14.441$                & $\underline{13.680}$    & $12.592$                    
                                                                    & \multirow{3}{*}{$\textbf{66}$}                                                       \\  
                                                        & MS-SSIM   & $0.440$                 & $0.390$                 & $0.316$                     
                                                                    & $0.187$                 & $0.180$                 & $\underline{0.182}$         
                                                                    & $0.221$                 & $0.227$                 & $\underline{0.225}$         
                                                                    & $0.280$                 & $0.264$                 & $0.240$                 \\  
                                                        & LPIPS     & $0.679$                 & $0.790$                 & $0.877$                     
                                                                    & $0.760$                 & $0.842$                 & $0.859$                     
                                                                    & $0.748$                 & $0.823$                 & $0.836$                     
                                                                    & $0.729$                 & $0.818$                 & $0.856$                 \\  
\cmidrule(ll){1-2}
\multirow{3}{*}{DMVFN~\cite{hu2023dynamic}}             & PSNR      & $\underline{19.605}$    & $\underline{17.708}$    & $\underline{17.228}$        
                                                                    & $11.821$                & $10.702$                & $10.613$                    
                                                                    & $\underline{13.445}$    & $12.071$                & $11.784$                    
                                                                    & $\underline{14.917}$    & $13.456$                & $\underline{13.170}$        
                                                                    & \multirow{3}{*}{$\underline{28}$}                                                       \\  
                                                        & MS-SSIM   & $\underline{0.487}$     & $\underline{0.430}$     & $\underline{0.418}$         
                                                                    & $\underline{0.241}$     & $\textbf{0.222}$        & $\textbf{0.223}$            
                                                                    & $\underline{0.317}$     & $\textbf{0.289}$        & $\textbf{0.277}$            
                                                                    & $\underline{0.347}$     & $\underline{0.313}$     & $\textbf{0.305}$        \\  
                                                        & LPIPS     & $\underline{0.379}$     & $0.629$                 & $0.697$                     
                                                                    & $0.604$                 & $0.840$                 & $0.880$                     
                                                                    & $\underline{0.564}$     & $0.790$                 & $0.855$                     
                                                                    & $\underline{0.516}$     & $0.753$                 & $0.811$                 \\  
\cmidrule(ll){1-2}
\multirow{3}{*}{SynSin~\cite{wiles2020synsin}}          & PSNR      & $17.876$                & $15.602$                & $14.635$                    
                                                                    & $11.839$                & $10.875$                & $10.639$                    
                                                                    & $12.915$                & $12.172$                & $11.896$                    
                                                                    & $14.175$                & $12.865$                & $12.378$                    
                                                                    & \multirow{3}{*}{$13$}                                                       \\  
                                                        & MS-SSIM   & $0.385$                 & $0.332$                 & $0.324$                     
                                                                    & $0.180$                 & $0.136$                 & $0.133$                     
                                                                    & $0.243$                 & $0.209$                 & $0.202$                     
                                                                    & $0.268$                 & $0.225$                 & $0.219$                 \\  
                                                        & LPIPS     & $0.580$                 & $0.643$                 & $0.672$                     
                                                                    & $0.703$                 & $0.763$                 & $0.781$                     
                                                                    & $0.689$                 & $0.730$                 & $0.748$                     
                                                                    & $0.658$                 & $0.712$                 & $0.734$                 \\  
\cmidrule(ll){1-2}
\multirow{3}{*}{Ours}                                   & PSNR      & $\textbf{21.811}$       & $\textbf{19.638}$       & $\textbf{18.832}$           
                                                                    & $\textbf{13.496}$       & $\underline{11.267}$    & $\underline{10.918}$        
                                                                    & $\textbf{13.999}$       & $\underline{12.392}$    & $\underline{12.203}$        
                                                                    & $\textbf{16.369}$       & $\textbf{14.377}$       & $\textbf{13.937}$           
                                                                    & \multirow{3}{*}{$13^{\dagger}$}                                             \\  
                                                        & MS-SSIM   & $\textbf{0.725}$        & $\textbf{0.571}$        & $\textbf{0.497}$            
                                                                    & $\textbf{0.461}$        & $\underline{0.220}$     & $0.161$                     
                                                                    & $\textbf{0.421}$        & $\underline{0.245}$     & $0.210$                     
                                                                    & $\textbf{0.532}$        & $\textbf{0.342}$        & $\underline{0.287}$     \\  
                                                        & LPIPS     & $\textbf{0.276}$        & $\textbf{0.366}$        & $\textbf{0.416}$            
                                                                    & $\textbf{0.450}$        & $\textbf{0.595}$        & $\textbf{0.632}$            
                                                                    & $\textbf{0.482}$        & $\textbf{0.580}$        & $\textbf{0.628}$            
                                                                    & $\textbf{0.404}$        & $\textbf{0.515}$        & $\textbf{0.560}$        \\  
\bottomrule
\end{tabular}
}
\begin{tablenotes}
\small
\item $^\dagger$Note that our ROS implementation runs DAv2 and Pulsar asynchronously, enabling a higher frame rate on real-time experiments.
\end{tablenotes}
\vspace{-10pt}
\end{table*}

\subsubsection*{\textbf{Depth Model Analysis}}

Recall that the purpose of learning our own local metric depth model is to allow our pipeline to work with robots without depth sensors, make up for low-bandwidth network properties that limit the reliability of receiving depth frames from the robot, and to generate complete depth images where standard sensors are noisy.
As such, to generate accurate future projections, it is important to learn a reliable depth model.
Leveraging DAv2, we first explore the feasibility of finetuning foundation models to our specific environment.
In Table~\ref{tab:depth_acc}, we report the average absolute relative error (AbsRel) and $\delta_1$ against known ground truth ZED values.
Across all three test growth stages, we outperform the pretrained DAv2 model significantly.

Tangentially, to evaluate the sensitivity of finetuned accuracy to the pretraining dataset and task loss, we further perform a comparative study by finetuning a DINOv2~\cite{oquab2024dinov2} model pretrained with sparse labels from the real-world KITTI dataset on the RDE task.
Impressively, we find DINOv2 outperforms DAv2-Terra in depth accuracy, even though it has been pretrained on a different task.
However, strictly looking at depth accuracy against ZED values gives a skewed understanding of the usability of the model for downstream rendering, since ZED does not provide labels for pixels that are outside of the camera's defined range.
As such, several sky pixels are not accounted for in the computation of depth accuracy.
To compare the quality of depth predictions for downstream rendering, we generate delayed reprojections of each test video using Pulsar ($\gamma=1\mathrm{e}{-5}$) with ground truth future poses.
Then, we compute the PSNR of valid projected pixels in each predicted image against the ground truth future image.
Here, we see that DAv2-Terra produces higher PSNR for early stage videos, where there are large patches of sky pixels, hinting that the finetuned DINOv2 model predicts sky depth inaccurately.

To test this hypothesis qualitatively, we visualize depth images from ZED, DINOv2, and DAv2, along with their future Pulsar renderings in Figure~\ref{fig:depth_comparison}.
Unsurprisingly, we see ZED reprojections align well with ground truth images, but have several holes (green).
Corroborating our hypothesis, we find DINOv2 generates inaccurate metric depth in sky regions, resulting in undesired warping in image generations.
Finally, the finetuned DAv2 model has some inaccuracies in depth images, but the errors are minor enough to result in a well-aligned reprojection prior to inpainting.
As such, considering the quality of renderings and the model's runtime speed, we choose DAv2-Terra as our pipeline's depth estimator.

\begin{figure}[t]
    \centering
    \includegraphics[width=\linewidth]{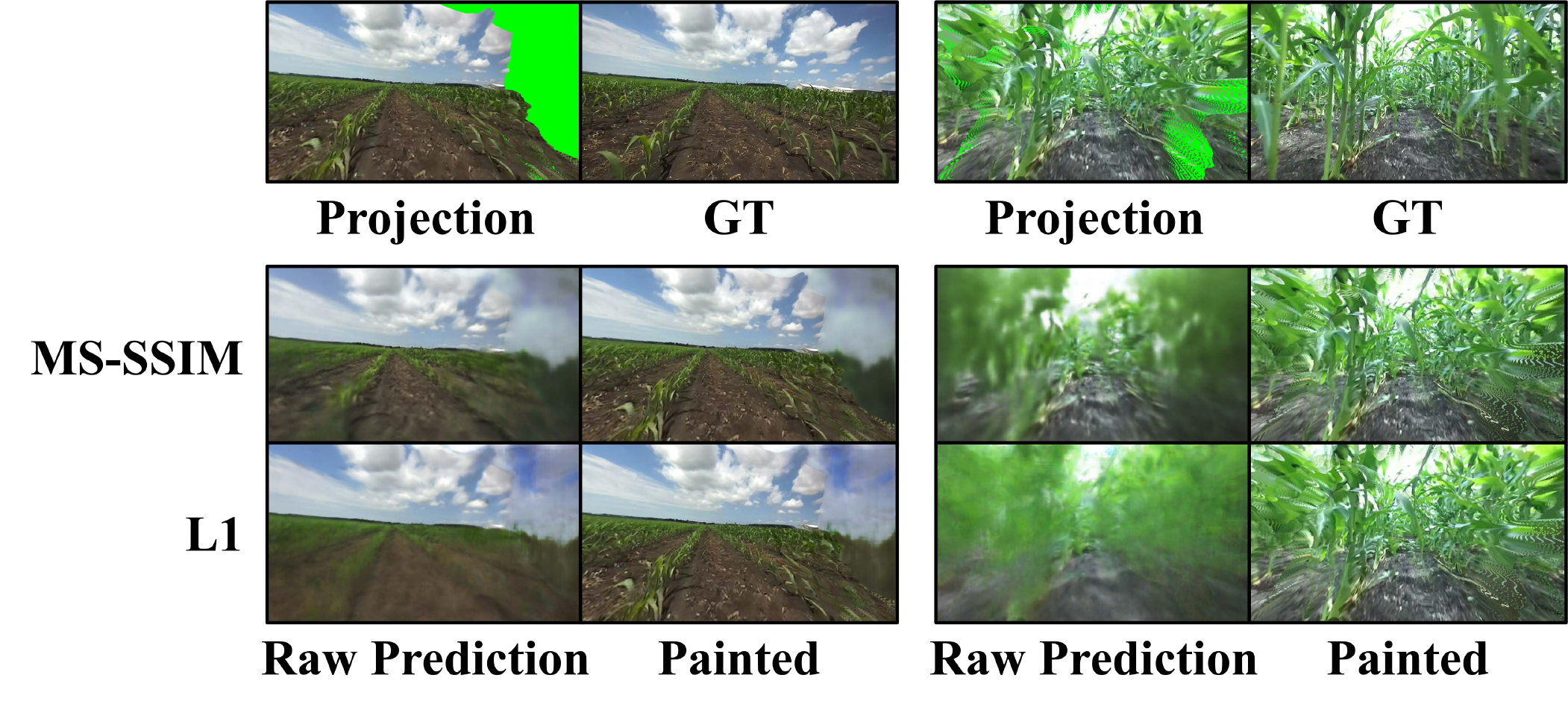}
    \vspace{-20pt}
    \caption{\textbf{Examples of ResNet inpainting model predictions.}}
    \label{fig:inpainting_comparison}
    \vspace{-15pt}
\end{figure}

\subsubsection*{\textbf{Inpainting Quality}}

With enough camera motion, any depth estimate used for reprojecting future frames with Pulsar will result in disocclusions in the rendered image.
To fill in these new holes, we train a ResNet-based model to predict cleaned versions of the future frame.
In Table~\ref{tab:inpainting}, we report the average quality of image generations from three different inpainting models on $5$- and $10$-timestep delayed test video feeds.
We first evaluate the model proposed by Telea~\cite{telea2004image}, which is used in the predictive display pipeline presented by Prakash~\etal~\cite{prakash2023predictive}.
While this approach achieves similar MS-SSIM to our ResNet models, it cannot run in real-time on our $8$ core Intel i$7$-$9700$ CPU.
Particularly, the iterative inpainting algorithm takes considerably longer time to predict pixels for images with larger and more disjoint holes.
While early stage images took on average $91$ ms to paint, each late stage reprojection took upwards of $5$ seconds.
In contrast, our ResNet model running on a 2080 GPU operates at $36$ FPS with $720$p input, and generates higher quality images.

As an ablation study, we train an additional inpainting model with the same architecture using MS-SSIM as its loss function, following recent findings from Shi~\etal~\cite{shi2022loss}, who report the superiority of structural similarity index measure-based losses for image generation.
Qualitatively, we find the raw predictions output by ResNet-MS-SSIM are sharper than the model trained with L1 loss in the regions without disocclusions in the reprojected image.
However, once we use the raw prediction to fill in holes from the reprojection, we see the painted regions contrast heavily with the remainder of the image.
On the other hand, while L1 loss leads to blurrier raw predictions, the predictions for disoccluded patches in the reprojected image blend in smoother, resulting in a higher PSNR.
Thus, we use the ResNet model trained with L1 loss as our inpainting model.
Examples of model predictions are provided in Figure~\ref{fig:inpainting_comparison}.

\balance 

\subsubsection*{\textbf{Comparison to Baselines}}

We present results on the accuracy of model predictions across each test video and different time delays in Table~\ref{tab:baseline_comparison} and example generations are provided in the supplentary video.
Generally, we find all methods perform best on early stage sequences compared to middle and late stage images due to fewer occlusions.
Similarly, as expected, larger delays lead to worse generations.
However, on average across the board, our method outperforms the others in all three metrics.
Particularly, C$+$S generates cropped and resized images which have high perceptual similarity to the original scene, but in fact align poorly with the true image.
Iterative approaches like SRVP and DMVFN incrementally produce more blurred, incomprehensible outputs as intermediate errors compound.
Finally, we find SynSin is unable to learn accurate enough depth or CNN features to decode future states accurately.
SRVP realizes the closest results to ours in late stage PSNR, but it requires training over five days on two NVIDIA A100 GPUs with $256\times256$ resolution images, whereas finetuning DAv2 and training our inpainting model each took one day on half the compute with the full resolution images.
DMVFN similarly achieves comparable MS-SSIM, but its blurry generations result in poor LPIPS.
However, it is worth noting that C$+$S, SynSin, and our model predict future frames assuming the world is static, resulting in inaccurate results when wind or the robot itself moves crops. 
We also find all methods perform poorly when conditioning generations on an occluded frame, leading us to develop real-time occlusion filters in future work.

\section{Real-Time Experiments}
\label{sec:online_eval_whole}

We also develop a ROS node to deploy our delay compensation method in real-time on TerraSentia rosbags.
A block diagram of the node is shown in Figure~\ref{fig:ros_block}.
To test the quality of the real-time compensated video feed, we emulate the conditions of varying network settings by skipping $5$ and $10$ frames (effectively requiring to compensate for $6$ and $3$ FPS videos respectively, from a $30$ Hz stream), and applying $250$ and $500$ ms delays to real-world rosbags in different growth stages on straight-path and turning trajectories.
Results are in the supplementary material and website.
We qualitatively find our model is capable of compensating for different frame rates and delays.
However, noisy odometry and poor predictions from our simplistic kinematic model lead to undesired jumps in the generated video under worse conditions.

\section{Conclusion}
\label{sec:conclusion}

We present an efficient and accurate modular learning-based pipeline for frame delay compensation in outdoor mobile robot teleoperation.
Future work includes integrating our ROS node into the real-world robot, developing controllers on the robot to compensate for delayed commands, and performing a large-scale user study to assess usability.


\section*{Acknowledgements}

N. C. thanks Jose Cuaran and Mateus Gasparino for help with the dataset, and Emerson Sie for developing the teleop network. 
This work was supported in part by the National Robotics Initiative 2.0 (NIFA\#2021-67021-33449) and AIFARMS through the Agriculture and Food Research Initiative (AFRI) grant no. 2020-67021-32799/project accession no.1024178 from the USDA/NIFA.
The robot platforms and data were provided by the Illinois Autonomous Farm and the Illinois Center for Digital Agriculture.
Additional resources were provided by the National Science Foundation’s (NSF) Major Research Instrumentation program, grant \#1725729.
Support for C. W. was provided by NSF proposal 2244580.


\bibliographystyle{IEEEtran}
\bibliography{IEEEabrv,IEEEexample,root}

\end{document}